\documentclass{article}

\usepackage{microtype}
\usepackage{subfigure}
\usepackage{url}
\usepackage{amsfonts}
\usepackage{amsmath}
\usepackage{amssymb}
\usepackage{amsthm}
\usepackage{mathrsfs}
\usepackage{tikz}
\usepackage{graphicx}
\usepackage{algorithm}
\usepackage{verbatim}
\usepackage{lmodern}
\usepackage{xcolor}
\usepackage[colorlinks=true]{hyperref}
\usepackage{booktabs}
\usepackage{multirow}
\usepackage{todonotes}
\usepackage{footnote}
\usepackage{wrapfig}
\usepackage{blindtext}
\usepackage[font=small]{caption}
\usepackage{mwe}
\makesavenoteenv{tabular}
\usepackage[toc,page]{appendix}

\newcommand{\cf}{\textit{c.f.}}
\newcommand{\eg}{\textit{e.g.}}
\newcommand{\ie}{\textit{i.e.}}

\newcommand{\x}{\mathbf{x}}

\newcommand{\m}{\mathbf{m}}

\newcommand{\h}{\mathbf{h}}
\newcommand{\W}{\mathbf{W}}
\renewcommand{\c}{\mathbf{c}}

\usepackage{comment}

\newcommand{\enas}{ENAS}

\usepackage{hyperref}

\usepackage[accepted]{icml2018}

\icmltitlerunning{Efficient Neural Architecture Search via Parameter Sharing}

\begin{document}

\twocolumn[
\icmltitle{Efficient Neural Architecture Search via Parameter Sharing}



\icmlsetsymbol{equal}{*}

\begin{icmlauthorlist}
\icmlauthor{Hieu Pham}{equal,googlebrain,cmu}
\icmlauthor{Melody Y. Guan}{equal,stanford}
\icmlauthor{Barret Zoph}{googlebrain}
\icmlauthor{Quoc V. Le}{googlebrain}
\icmlauthor{Jeff Dean}{googlebrain}
\end{icmlauthorlist}

\icmlaffiliation{cmu}{Language Technology Institute, Carnegie Mellon University}
\icmlaffiliation{googlebrain}{Google Brain}
\icmlaffiliation{stanford}{Department of Computer Science, Stanford University}

\icmlcorrespondingauthor{Hieu Pham}{hyhieu@cmu.edu}
\icmlcorrespondingauthor{Melody Y. Guan}{mguan@stanford.edu}

\icmlkeywords{Neural Architecture Search, Parameter Sharing, Stochastic Computation}

\vskip 0.3in
]

\printAffiliationsAndNotice{\icmlEqualContribution}

\begin{abstract}
  We propose \textit{Efficient Neural Architecture Search} (\enas), a fast and inexpensive approach for automatic model design. In \enas, a controller discovers neural network architectures by searching for an optimal subgraph within a large computational graph. The controller is trained with policy gradient to select a subgraph that maximizes the expected reward on a validation set. Meanwhile the model corresponding to the selected subgraph is trained to minimize a canonical cross entropy loss. Sharing parameters among child models allows \enas~to deliver strong empirical performances, while using much fewer GPU-hours than existing automatic model design approaches, and notably, 1000x less expensive than standard Neural Architecture Search. On the Penn Treebank dataset, \enas~discovers a novel architecture that achieves a test perplexity of $55.8$, establishing a new state-of-the-art among all methods without post-training processing. On the CIFAR-10 dataset, \enas~finds a novel architecture that achieves $2.89\%$ test error, which is on par with the $2.65\%$ test error of NASNet~\citep{nas_module}.
\end{abstract}

\vspace{-0.8cm}
\section{\label{sec:intro}Introduction}

Neural architecture search (NAS) has been successfully applied to design model architectures for image classification and language models~\citep{nas,nas_module,nas_by_transformation,pnas,hierarchical_nas}. In NAS, an RNN controller is trained in a loop: the controller first samples a candidate architecture, \ie~a \textit{child model}, and then trains it to convergence to measure its performance on the task of desire. The controller then uses the performance as a guiding signal to find more promising architectures. This process is repeated for many iterations. Despite its impressive empirical performance, NAS is computationally expensive and time consuming, \eg~\citet{nas_module} use $450$ GPUs for $3$-$4$ days (\ie~32,400-43,200 GPU hours). Meanwhile, using less resources tends to produce less compelling results~\citep{deep_arc,nas_dqn}. We observe that the computational bottleneck of NAS is the training of each child model to convergence, only to measure its accuracy whilst throwing away all the trained weights.

The main contribution of this work is to improve the efficiency of NAS by \textit{forcing all child models to share weights} to eschew training each child model from scratch to convergence. The idea has apparent complications, as different child models might utilize their weights differently, but was encouraged by previous work on transfer learning and multitask learning, which established that parameters learned for a particular model on a particular task can be used for other models on other tasks, with little to no modifications~\citep{conv_features_transfer,nmt_transfer,multitask_seq2seq}.

We empirically show that not only is sharing parameters among child models possible, but it also allows for very strong performance. Specifically, on CIFAR-10, our method achieves a test error of $2.89\%$, compared to $2.65\%$ by NAS. On Penn Treebank, our method achieves a test perplexity of $55.8$, which significantly outperforms NAS's test perplexity of $62.4$~\citep{nas} and which is a new state-of-the-art among Penn Treebank's approaches that do not utilize post-training processing. Importantly, in all of our experiments, for which we use a single Nvidia GTX 1080Ti GPU, the search for architectures takes less than $16$ hours. Compared to NAS, this is a reduction of GPU-hours by more than 1000x. Due to its efficiency, we name our method \textit{Efficient Neural Architecture Search} (\enas).
\section{\label{sec:enas}Methods}
\begin{figure*}[htb!]
  \centering
  \includegraphics[width=0.9\textwidth]{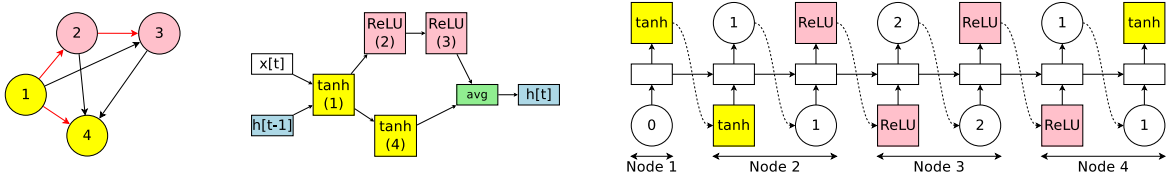}
  \caption{\label{fig:enas_rnn_example}An example of a recurrent cell in our search space with $4$ computational nodes. \textit{Left:} The computational DAG that corresponds to the recurrent cell. The red edges represent the flow of information in the graph. \textit{Middle:} The recurrent cell. \textit{Right:} The outputs of the controller RNN that result in the cell in the middle and the DAG on the left. Note that nodes $3$ and $4$ are never sampled by the RNN, so their results are averaged and are treated as the cell's output.}
\end{figure*}

\begin{figure}[htb!]
  \centering
  \includegraphics[width=0.3\textwidth]{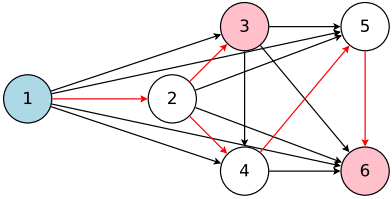}
  \caption{\label{fig:enas_dag}The graph represents the entire search space while the red arrows define a model in the search space, which is decided by a controller. Here, node $1$ is the input to the model whereas nodes $3$ and $6$ are the model's outputs.}
\end{figure}

Central to the idea of \enas~is the observation that all of the graphs which NAS ends up iterating over can be viewed as sub-graphs of a larger graph. In other words, we can represent NAS's search space using a \textit{single} directed acyclic graph (DAG). Figure~\ref{fig:enas_dag} illustrates a generic example DAG, where an architecture can be realized by taking a subgraph of the DAG. Intuitively, \enas's DAG is the superposition of all possible child models in a search space of NAS, where the nodes represent the local computations and the edges represent the flow of information. The local computations at each node have their own parameters, which are used only when the particular computation is activated. Therefore, \enas's design allows parameters to be shared among all child models, \ie~architectures, in the search space.

In the following, we facilitate the discussion of \enas~with an example that illustrates how to design a cell for recurrent neural networks from a specified DAG and a controller (Section~\ref{sec:enas_rnn}). We will then explain how to train \enas~and how to derive architectures from \enas's controller (Section~\ref{sec:model_policy}). Finally, we will explain our search space for designing convolutional architectures (Sections~\ref{sec:cifar10_macro} and~\ref{sec:cifar10_micro}).

\subsection{\label{sec:enas_rnn}Designing Recurrent Cells}
To design recurrent cells, we employ a DAG with $N$ nodes, where the nodes represent local computations, and the edges represent the flow of information between the $N$ nodes. \enas's controller is an RNN that decides: 1) which edges are activated and 2) which computations are performed at each node in the DAG. This design of our search space for RNN cells is different from the search space for RNN cells in~\citet{nas}, where the authors fix the topology of their architectures as a binary tree and only learn the operations at each node of the tree. In contrast, our search space allows \enas~to design both the topology and the operations in RNN cells, and hence is more flexible.

To create a recurrent cell, the controller RNN samples $N$ blocks of decisions. Here we illustrate the \enas~mechanism via a simple example recurrent cell with $N = 4$ computational nodes (visualized in Figure~\ref{fig:enas_rnn_example}). Let $\x_t$ be the input signal for a recurrent cell (\eg~word embedding), and $\h_{t-1}$ be the output from the previous time step. We sample as follows.
\begin{enumerate}
  \small
  \item At node 1: The controller first samples an activation function. In our example, the controller chooses the $\tanh$ activation function, which means that node $1$ of the recurrent cell should compute $h_1 = \tanh{(\x_t \cdot \W^{(\x)} + \h_{t-1} \cdot \W^{(\h)}_1)}$.
  \item At node 2: The controller then samples a previous index and an activation function. In our example, it chooses the previous index $1$ and the activation function $\text{ReLU}$. Thus, node $2$ of the cell computes $h_2 = \text{ReLU}(h_1 \cdot \W^{(\h)}_{2, 1})$.
  \item At node 3: The controller again samples a previous index and an activation function. In our example, it chooses the previous index $2$ and the activation function $\text{ReLU}$. Therefore, $h_3 = \text{ReLU}(h_2 \cdot \W^{(\h)}_{3, 2})$.
  \item At node 4: The controller again samples a previous index and an activation function. In our example, it chooses the previous index $1$ and the activation function $\tanh$, leading to $h_4 = \tanh{(h_1 \cdot \W^{(\h)}_{4, 1})}$.
  \item For the output, we simply average all the loose ends, \ie~the nodes that are not selected as inputs to any other nodes. In our example, since the indices $3$ and $4$ were never sampled to be the input for any node, the recurrent cell uses their average $(h_3 + h_4) / 2$ as its output. In other words, $\h_t = (h_3 + h_4) / 2$.
\end{enumerate}

In the example above, we note that for each pair of nodes $j < \ell$, there is an independent parameter matrix $\W^{(\h)}_{\ell, j}$. As shown in the example, by choosing the previous indices, the controller also decides which parameter matrices are used. Therefore, in \enas, all recurrent cells in a search space share the same set of parameters.

Our search space includes an exponential number of configurations. Specifically, if the recurrent cell has $N$ nodes and we allow $4$ activation functions (namely $\tanh$, $\text{ReLU}$, $\text{identity}$, and $\text{sigmoid}$), then the search space has $4^{N} \times N!$ configurations. In our experiments, $N = 12$, which means there are approximately $10^{15}$ models in our search space.

\subsection{\label{sec:model_policy}Training \enas~and Deriving Architectures}
Our controller network is an LSTM with $100$ hidden units~\citep{lstm}. This LSTM samples decisions via softmax classifiers, in an autoregressive fashion: the decision in the previous step is fed as input embedding into the next step. At the first step, the controller network receives an empty embedding as input.

In \enas, there are two sets of learnable parameters: the parameters of the controller LSTM, denoted by $\theta$, and the shared parameters of the child models, denoted by $\omega$. The training procedure of \enas~consists of two interleaving phases. The first phase trains $\omega$, the shared parameters of the child models, on a whole pass through the training data set. For our Penn Treebank experiments, $\omega$ is trained for about $400$ steps, each on a minibatch of $64$ examples, where the gradient $\nabla_\omega$ is computed using back-propagation through time, truncated at $35$ time steps. Meanwhile, for CIFAR-10, $\omega$ is trained on $45,000$ training images, separated into minibatches of size $128$, where $\nabla_\omega$ is computed using standard back-propagation. The second phase trains $\theta$, the parameters of the controller LSTM, for a fixed number of steps, typically set to $2000$ in our experiments. These two phases are alternated during the training of ENAS. More details are as follows.

\paragraph{Training the shared parameters $\omega$ of the child models.} In this step, we fix the controller's policy $\pi(\m; \theta)$ and perform stochastic gradient descent (SGD) on $\omega$ to minimize the expected loss function $\mathbb{E}_{\m \sim \pi} \left[ \mathcal{L}(\m; \omega) \right]$. Here, $\mathcal{L}(\m; \omega)$ is the standard cross-entropy loss, computed on a minibatch of training data, with a model $\m$ sampled from $\pi(\m; \theta)$. The gradient is computed using the Monte Carlo estimate
\begin{align}
  \label{eqn:update_omega}
  \nabla_\omega \mathbb{E}_{\m \sim \pi(\m; \theta)}
  \left[ \mathcal{L}(\m; \omega) \right]
    &\approx \frac{1}{M} \sum_{i=1}^{M} \nabla_\omega \mathcal{L}(\m_i, \omega),
\end{align}
where $\m_i$'s are sampled from $\pi(\m; \theta)$ as described above. Eqn~\ref{eqn:update_omega} provides an unbiased estimate of the gradient $\nabla_\omega \mathbb{E}_{\m \sim \pi(\m; \theta)} \left[ \mathcal{L}(\m; \omega) \right]$. However, this estimate has a higher variance than the standard SGD gradient, where $\m$ is fixed. Nevertheless -- and this is perhaps surprising -- we find that $M = 1$ works just fine, \ie~we can update $\omega$ using the gradient from \textit{any single model} $\m$ sampled from $\pi(\m; \theta)$. As mentioned, we train $\omega$ during a entire pass through the training data.

\paragraph{Training the controller parameters $\theta$.} In this step, we fix $\omega$ and update the policy parameters $\theta$, aiming to maximize the expected reward $\mathbb{E}_{\m \sim \pi(\m; \theta)}\left[ \mathcal{R}(\m, \omega) \right]$. We employ the Adam optimizer~\citep{adam}, for which the gradient is computed using REINFORCE~\citep{reinforce}, with a moving average baseline to reduce variance.

The reward $\mathcal{R}(\m, \omega)$ is computed on \textit{the validation set}, rather than on the training set, to encourage \enas~to select models that generalize well rather than models that overfit the training set well. In our language model experiment, the reward function is $c/\text{valid\_ppl}$, where the perplexity is computed on a minibatch of validation data. In our image classification experiments, the reward function is the accuracy on a minibatch of validation images.

\paragraph{Deriving Architectures.} We discuss how to derive novel architectures from a trained \enas~model. We first sample several models from the trained policy $\pi(\m, \theta)$. For each sampled model, we compute its reward on \textit{a single minibatch} sampled from the validation set. We then take only the model with the highest reward to re-train from scratch. It is possible to improve our experimental results by training all the sampled models from scratch and selecting the model with the highest performance on a separated validation set, as done by other works~\citep{nas,nas_module,pnas,hierarchical_nas}. However, our method yields similar performance whilst being much more economical.

\subsection{\label{sec:cifar10_macro}Designing Convolutional Networks}
\begin{figure}[htb!]
  \centering
  \includegraphics[width=0.45\textwidth]{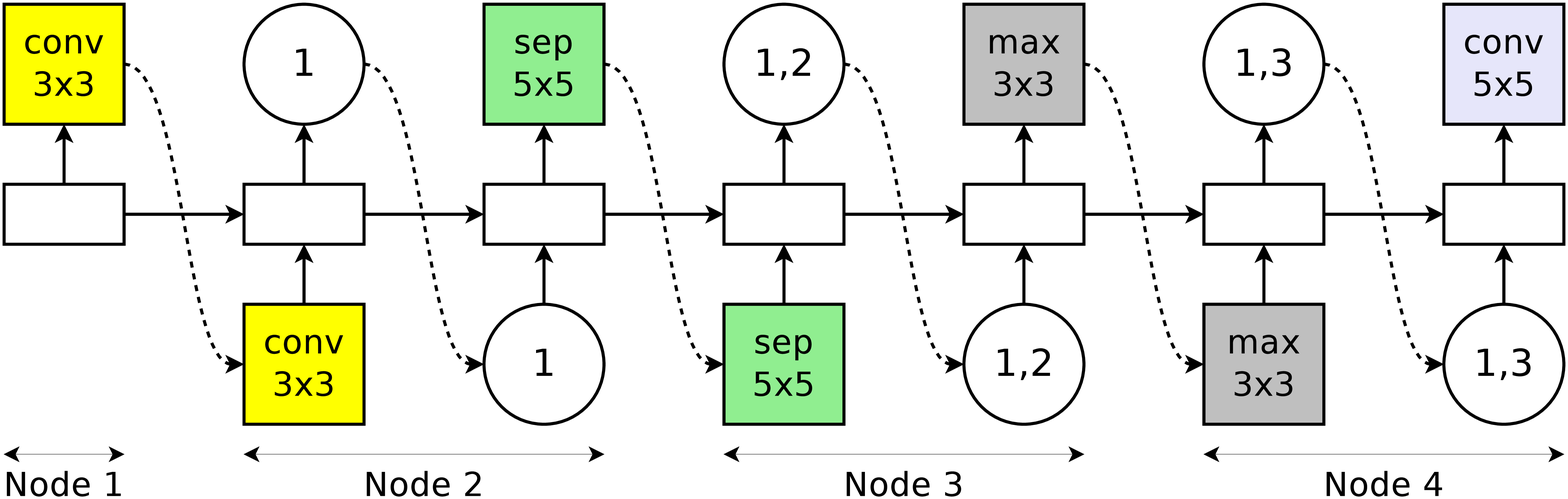}
  \includegraphics[height=0.45\textwidth,angle=270]{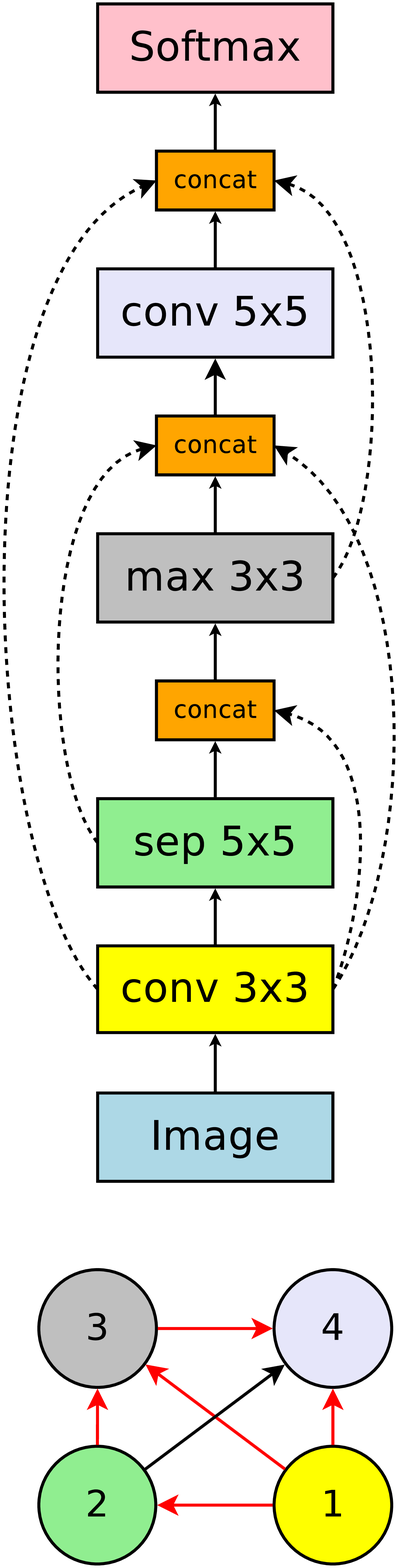}
  \caption{\label{fig:enas_conv_example}An example run of a recurrent cell in our search space with $4$ computational nodes, which represent $4$ layers in a convolutional network. \textit{Top:} The output of the controller RNN. \textit{Bottom Left:} The computational DAG corresponding to the network's architecture. Red arrows denote the active computational paths. \textit{Bottom Right:} The complete network. Dotted arrows denote skip connections.}
\end{figure}
We now discuss the search space for convolutional architectures. Recall that in the search space of the recurrent cell, the controller RNN samples two decisions at each decision block: 1) what previous node to connect to and 2) what activation function to use. In the search space for convolutional models, the controller RNN also samples two sets of decisions at each decision block: 1) what previous nodes to connect to and 2) what computation operation to use. These decisions construct a layer in the convolutional model.

The decision of what previous nodes to connect to allows the model to form skip connections~\citep{res_net,nas}. Specifically, at layer $k$, up to $k-1$ mutually distinct previous indices are sampled, leading to $2^{k-1}$ possible decisions at layer $k$. We provide an illustrative example of sampling a convolutional network in Figure~\ref{fig:enas_conv_example}. In this example, at layer $k=4$, the controller samples previous indices $\{1, 3\}$, so the outputs of layers $1$ and $3$ are concatenated along their depth dimension and sent to layer $4$.

Meanwhile, the decision of what computation operation to use sets a particular layer into convolution or average pooling or max pooing. The $6$ operations available for the controller are: convolutions with filter sizes $3 \times 3$ and $5 \times 5$, depthwise-separable convolutions with filter sizes $3 \times 3$ and $5 \times 5$~\citep{xception}, and max pooling and average pooling of kernel size $3 \times 3$. As for recurrent cells, each operation at each layer in our \enas~convolutional network has a distinct set of parameters.

Making the described set of decisions for a total of $L$ times, we can sample a network of $L$ layers. Since all decisions are independent, there are $6^{L} \times 2^{L(L-1) / 2}$ networks in the search space. In our experiments, $L = 12$, resulting in $1.6 \times 10^{29}$ possible networks.

\subsection{\label{sec:cifar10_micro}Designing Convolutional Cells}
\begin{figure}[htb!]
  \centering
  \includegraphics[height=0.45\textwidth,angle=270]{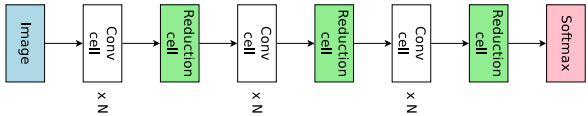}
  \caption{\label{fig:micro_search_space}Connecting $3$ blocks, each with $N$ convolution cells and $1$ reduction cell, to make the final network.}
\end{figure}

Rather than designing the entire convolutional network, one can design smaller modules and then connect them together to form a network~\citep{nas_module}. Figure~\ref{fig:micro_search_space} illustrates this design, where the convolutional cell and reduction cell architectures are to be designed. We now discuss how to use \enas~to search for the architectures of these cells.

We utilize the \enas~computational DAG with $B$ nodes to represent the computations that happen \textit{locally in a cell}. In this DAG, node $1$ and node $2$ are treated as the cell's inputs, which are the outputs of the two previous cells in the final network (see Figure~\ref{fig:micro_search_space}). For each of the remaining $B-2$ nodes, we ask the controller RNN to make two sets of decisions: 1) two previous nodes to be used as inputs to the current node and 2) two operations to apply to the two sampled nodes. The $5$ available operations are: identity, separable convolution with kernel size $3 \times 3$ and $5 \times 5$, and average pooling and max pooling with kernel size $3 \times 3$. At each node, after the previous nodes and their corresponding operations are sampled, the operations are applied on the previous nodes, and their results are added.

\begin{figure}[htb!]
  \centering
  \includegraphics[width=0.45\textwidth]{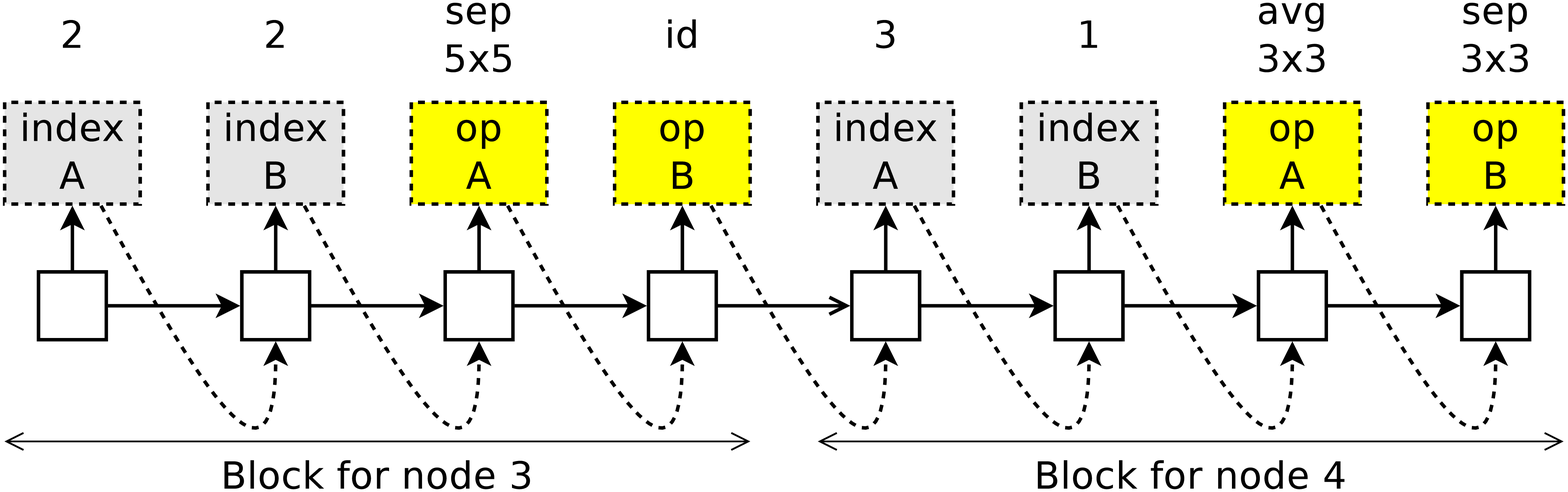}\\~~
  \includegraphics[width=0.12\textwidth]{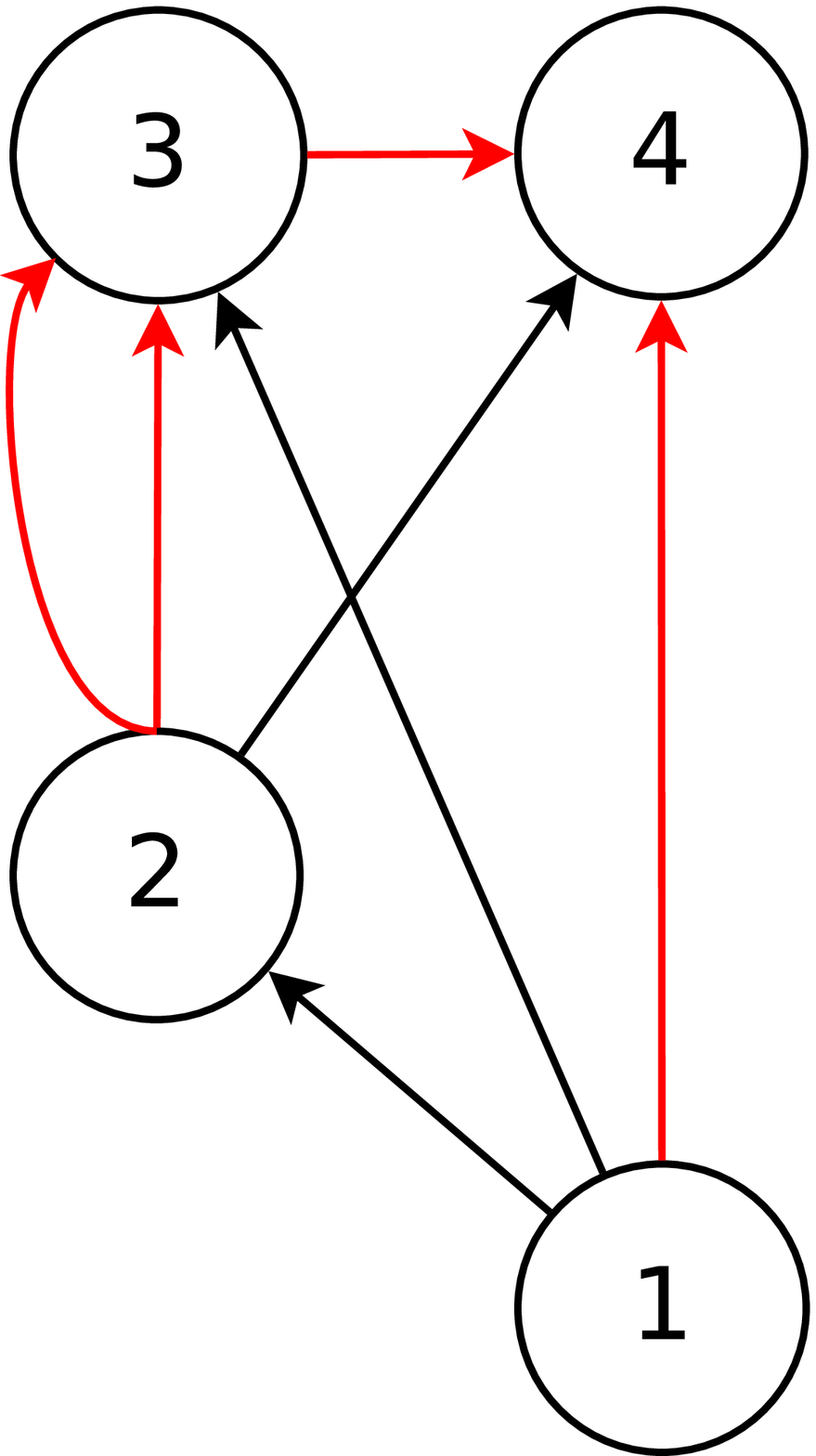}~~~\includegraphics[height=0.4\textwidth]{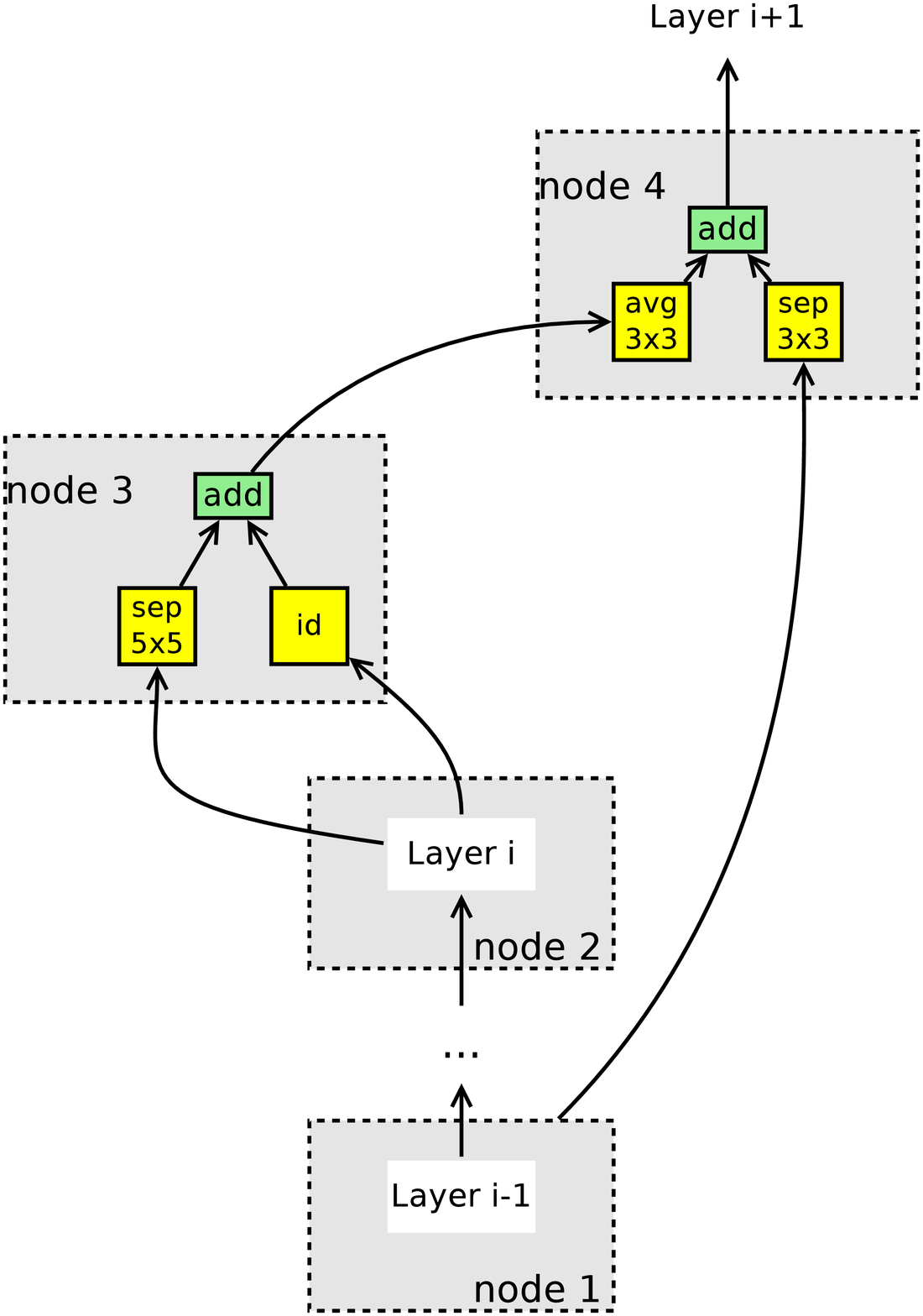}
  \caption{\label{fig:cifar10_micro_example}An example run of the controller for our search space over convolutional cells. \textit{Top:} the controller's outputs. In our search space for convolutional cells, node $1$ and node $2$ are the cell's inputs, so the controller only has to design node $3$ and node $4$. \textit{Bottom Left:} The corresponding DAG, where red edges represent the activated connections. \textit{Bottom Right:} the convolutional cell according to the controller's sample.}
\end{figure}
As before, we illustrate the mechanism of our search space with an example, here with $B=4$ nodes (refer to Figure~\ref{fig:cifar10_micro_example}). Details are as follows.
\begin{enumerate}
  \small
  \item Nodes $1$, $2$ are input nodes, so no decisions are needed for them. Let $h_1$, $h_2$ be the outputs of these nodes.
  \item At node 3: the controller samples two previous nodes and two operations. In Figure~\ref{fig:cifar10_micro_example} \textit{Top Left}, it samples \textit{node 2}, \textit{node 2}, \textit{separable\_conv\_5x5}, and \textit{identity}. This means that $h_3 = \text{sep\_conv\_5x5}(h_2) + \text{id}(h_2)$.
  \item At node 4: the controller samples \textit{node 3}, \textit{node 1}, \textit{avg\_pool\_3x3}, and \textit{sep\_conv\_3x3}. This means that $h_4 = \text{avg\_pool\_3x3}(h_3) + \text{sep\_conv\_3x3}(h_1)$.
  \item Since all nodes but $h_4$ were used as inputs to at least another node, the only loose end, $h_4$, is treated as the cell's output. If there are multiple loose ends, they will be concatenated along the depth dimension to form the cell's output.
\end{enumerate}
A reduction cell can also be realized from the search space we discussed, simply by: 1) sampling a computational graph from the search space, and 2) applying all operations with a stride of $2$. A reduction cell thus reduces the spatial dimensions of its input by a factor of $2$. Following~\citet{nas_module}, we sample the reduction cell conditioned on the convolutional cell, hence making the controller RNN run for a total of $2(B-2)$ blocks.

\begin{table*}[t!]
  \centering
  \begin{minipage}{1.0\textwidth}
    \small
    \centering
    \begin{tabular}{l|l|cc}
    \toprule
      \multirow{2}{*}{\textbf{Architecture}} & \multirow{2}{*}{\textbf{Additional Techniques}}
        & \textbf{Params} & \textbf{Test} \\
      & & (million)       & \textbf{PPL} \\
    \midrule
      LSTM~\citep{rnn_regularization}  & Vanilla Dropout & 66 & 78.4 \\
      LSTM~\citep{variational_dropout_rnn} & VD & 66 & 75.2 \\
      LSTM~\citep{lstm_weight_tying} & VD, WT & 51 & 68.5 \\
      LSTM~\citep{ptb_hparams} & Hyper-parameters Search & 24 & 59.5 \\
      LSTM~\citep{mos} & VD, WT, $\ell_2$, AWD, MoC & 22 & 57.6 \\
      LSTM~\citep{awd_lstm} & VD, WT, $\ell_2$, AWD & 24 & 57.3 \\
      LSTM~\citep{mos} & VD, WT, $\ell_2$, AWD, MoS & \textbf{22} & \textbf{56.0} \\
    \midrule
      RHN~\citep{rhn} & VD, WT & 24 & 66.0 \\
    \midrule
      NAS~\citep{nas}   & VD, WT & 54 & 62.4 \\
    \midrule
      \enas~      & VD, WT, $\ell_2$ & \textbf{24} & \textbf{55.8} \\
    \bottomrule
    \end{tabular}
    \caption{\label{tab:ptb_results}Test perplexity on Penn Treebank of \enas~and other baselines. Abbreviations: RHN is \textit{Recurrent Highway Network}, VD is \textit{Variational Dropout}; WT is \textit{Weight Tying}; $\ell_2$ is \textit{Weight Penalty}; AWD is \textit{Averaged Weight Drop}; MoC is \textit{Mixture of Contexts}; MoS is \textit{Mixture of Softmaxes}.}
  \end{minipage}
\end{table*}

Finally, we estimate the complexity of this search space. At node $i$ ($3 \leq i \leq B$), the controller can select any two nodes from the $i - 1$ previous nodes, and any two operations from $5$ operations. As all decisions are independent, there are $(5 \times (B-2)!)^2$ possible cells. Since we independently sample for a convolutional cell and a reduction cell, the final size of the search space is $(5 \times (B-2)!)^4$. With $B = 7$ as in our experiments, the search space can realize $1.3 \times 10^{11}$ final networks, making it significantly smaller than the search space for entire convolutional networks (Section~\ref{sec:cifar10_macro}).
\section{\label{sec:exp}Experiments}
We first present our experimental results from employing \enas~to design recurrent cells on the Penn Treebank dataset and convolutional architectures on the CIFAR-10 dataset. We then present an ablation study which asserts the role of \enas~in discovering novel architectures.

\subsection{\label{sec:ptb}Language Model with Penn Treebank}
\paragraph{Dataset and Settings.} Penn Treebank~\citep{penntreebank} is a well-studied benchmark for language model. We use the standard pre-processed version of the dataset, which is also used by previous works, \eg~\citet{rnn_regularization}.

Since the goal of our work is to discover cell architectures, we only employ the standard training and test process on Penn Treebank, and do not utilize post-training techniques such as neural cache~\citep{neural_cache} and dynamic evaluation~\citep{ptb_dynamic_eval}. Additionally, as~\citet{rnn_params_count} have established that RNN models with more parameters can learn to store more information, we limit the size of our \enas~cell to $24M$ parameters. We also do not tune our hyper-parameters extensively like~\citet{ptb_hparams}, nor do we train multiple architectures and select the best one based on their validation perplexities like~\citet{nas}. Therefore, \enas~is not at any advantage, compared to~\citet{nas,mos,ptb_hparams}, and its improved performance is only due to the cell's architecture.

\paragraph{Training details.} Our controller is trained using Adam, with a learning rate of $0.00035$. To prevent premature convergence, we also use a tanh constant of $2.5$ and a temperature of $5.0$ for the sampling logits~\citep{neural_combi,neural_optimizer_search}, and add the controller's sample entropy to the reward, weighted by $0.0001$. Additionally, we augment the simple transformations between nodes in the constructed recurrent cell with highway connections~\citep{rhn}. For instance, instead of having $h_2 =\text{ReLU}(h_1 \cdot \W^{(\h)}_{2, 1})$ as shown in the example from Section~\ref{sec:enas_rnn}, we have $h_2 = c_2 \otimes \text{ReLU}(h_1 \cdot \W^{(\h)}_{2, 1}) + (1 - c_2) \otimes h_1$, where $c_2 = \text{sigmoid}(h_1 \cdot \W^{(\c)}_{2, 1})$ and $\otimes$ denotes elementwise multiplication.

The shared parameters of the child models $\omega$ are trained using SGD with a learning rate of $20.0$, decayed by a factor of $0.96$ after every epoch starting at epoch $15$, for a total of $150$ epochs. We clip the norm of the gradient $\nabla_\omega$ at $0.25$. We find that using a large learning rate whilst clipping the gradient norm at a small threshold makes the updates on $\omega$ more stable.
We utilize three regularization techniques on $\omega$: an $\ell_2$ regularization weighted by $10^{-7}$; variational dropout~\citep{variational_dropout_rnn}; and tying word embeddings and softmax weights~\citep{lstm_weight_tying}. More details are in Appendix~\ref{sec:ptb_details}.

\paragraph{Results.} Running on a single Nvidia GTX 1080Ti GPU, \enas~finds a recurrent cell in about $10$ hours. In Table~\ref{tab:ptb_results}, we present the performance of the \enas~cell as well as other baselines that do not employ post-training processing. The \enas~cell achieves a test perplexity of $55.8$, which is on par with the existing state-of-the-art of $56.0$ achieved by \textit{Mixture of Softmaxes} (MoS)~\citep{mos}. Note that we do not apply MoS to the \enas~cell. Importantly, \enas~cell outperforms NAS~\citep{nas} by more than $6$ perplexity points, whilst the search process of \enas, in terms of GPU hours, is more than $1000$x faster.

\begin{figure}[h!]
  \centering
  \includegraphics[height=0.45\textwidth,angle=270]{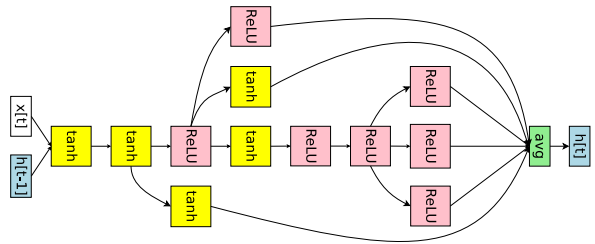}
  \caption{\label{fig:ptb_enas_cell}The RNN cell \enas~discovered for Penn Treebank.}
\end{figure}

Our \enas~cell, visualized in Figure~\ref{fig:ptb_enas_cell}, has a few interesting properties. First, all non-linearities in the cell are either ReLU or tanh, even though the search space also has two other functions: identity and sigmoid. Second, we suspect this cell is a local optimum, similar to the observations made by~\citet{nas}. When we randomly pick some nodes and switch the non-linearity into identity or sigmoid, the perplexity increases up to $8$ points. Similarly, when we randomly switch some ReLU nodes into tanh or vice versa, the perplexity also increases, but only up to $3$ points. Third, as shown in Figure~\ref{fig:ptb_enas_cell}, the output of our \enas~cell is an average of $6$ nodes. This behavior is similar to that of \textit{Mixture of Contexts} (MoC)~\citep{mos}. Not only does \enas~independently discover MoC, but it also learns to balance between i) the number of contexts to mix, which increases the model's expressiveness, and ii) the depth of the recurrent cell, which learns more complex transformations~\citep{rhn}.

\subsection{\label{sec:cifar10}Image Classification on CIFAR-10}
\paragraph{Dataset.} The CIFAR-10 dataset~\citep{cifar10} consists of $50,000$ training images and $10,000$ test images. We use the standard data pre-processing and augmentation techniques, \ie~subtracting the channel mean and dividing the channel standard deviation, centrally padding the training images to $40 \times 40$ and randomly cropping them back to $32 \times 32$, and randomly flipping them horizontally.

\paragraph{Search spaces.} We apply~\enas~to two search spaces: 1) the~\textit{macro search space} over entire convolutional models (Section~\ref{sec:cifar10_macro}); and 2) the~\textit{micro search space} over convolutional cells (Section~\ref{sec:cifar10_micro}).

\paragraph{Training details.} The shared parameters $\omega$ are trained with Nesterov momentum~\citep{nesterov}, where the learning rate follows the cosine schedule with $l_{\text{max}} = 0.05$, $\l_{\text{min}} = 0.001$, $T_0 = 10$, and $T_{\text{mul}} = 2$~\citep{cosine_lr}. Each architecture search is run for $310$ epochs. We initialize $\omega$ with He initialization~\citep{he_init}. We also apply an $\ell_2$ weight decay of $10^{-4}$. We train the architectures recommended by the controller using the same settings.

The policy parameters $\theta$ are initialized uniformly in $[-0.1, 0.1]$, and trained with Adam at a learning rate of $0.00035$. Similar to the procedure in Section~\ref{sec:ptb}, we apply a tanh constant of $2.5$ and a temperature of $5.0$ to the controller's logits, and add the controller entropy to the reward, weighted by $0.1$. Additionally, in the macro search space, we enforce the sparsity in the skip connections by adding to the reward the KL divergence between: 1) the skip connection probability between any two layers and 2) our chosen probability $\rho = 0.4$, which represents the prior belief of a skip connection being formed. This KL divergence term is weighted by $0.8$. More training details are in Appendix~\ref{sec:cifar10_details}.

\begin{table*}[htb!]
  \centering
  \begin{minipage}{1.\textwidth}
    \centering
    \small
    \begin{tabular}{lcccc}
    \toprule
      \multirow{2}{*}{\textbf{Method}} & \multirow{2}{*}{\textbf{GPUs}} & \textbf{Times} & \textbf{Params} & \textbf{Error} \\
      & & (days) & (million) & ($\%$) \\
    \midrule
      DenseNet-BC~\citep{dense_net} & $-$ & $-$ & 25.6 & 3.46 \\
      DenseNet + Shake-Shake~\citep{shake_shake} & $-$ & $-$ & 26.2 & 2.86 \\
      DenseNet + CutOut~\citep{cut_out} & $-$ & $-$ & 26.2 & \textbf{2.56} \\
    \midrule \midrule
      Budgeted Super Nets~\citep{super_network} & $-$ & $-$ & $-$ & 9.21 \\
      ConvFabrics~\citep{conv_fabrics} & $-$ & $-$ & 21.2 & 7.43 \\
      Macro NAS + Q-Learning~\citep{nas_dqn} & 10 & 8-10 & 11.2 & 6.92 \\
      Net Transformation~\citep{nas_by_transformation} & 5 & 2 & 19.7 & 5.70 \\
      FractalNet~\citep{fractal_net} & $-$ & $-$ & 38.6 & 4.60 \\
      SMASH~\citep{smash_net} & 1 & 1.5 & 16.0 & 4.03 \\
      NAS~\citep{nas} & 800 & 21-28 & 7.1 & 4.47 \\
      NAS + more filters~\citep{nas} & 800 & 21-28 & 37.4 & \textbf{3.65} \\
    \midrule
      \enas~+ macro search space & 1 & 0.32 & 21.3 & 4.23 \\  
      \enas~+ macro search space + more channels & 1 & 0.32 & 38.0 & \textbf{3.87} \\
    \midrule \midrule
      Hierarchical NAS~\citep{hierarchical_nas} & 200 & 1.5 & 61.3 & 3.63 \\
      Micro NAS + Q-Learning~\citep{nas_module_dqn} & 32 & 3 & $-$ & 3.60 \\
      Progressive NAS~\citep{pnas} & 100 & 1.5 & 3.2 & 3.63 \\
      NASNet-A~\citep{nas_module} &  450 & 3-4 & 3.3 & 3.41 \\
      NASNet-A + CutOut~\citep{nas_module} & 450 & 3-4 & 3.3 & \textbf{2.65} \\
    \midrule
      \enas~+ micro search space & 1 & 0.45 & 4.6 & 3.54 \\
      \enas~+ micro search space + CutOut & 1 & 0.45 & 4.6 & \textbf{2.89} \\
    \bottomrule
    \end{tabular}
    \caption{\label{tab:cifar10_results}Classification errors of \enas~and baselines on CIFAR-10. In this table, the first block presents DenseNet, one of the state-of-the-art architectures designed by human experts. The second block presents approaches that design the entire network. The last block presents techniques that design modular cells which are combined to build the final network.}
  \end{minipage}
\end{table*}

\begin{figure*}[htb!]
  \centering
  \includegraphics[height=0.75\textwidth,angle=270]{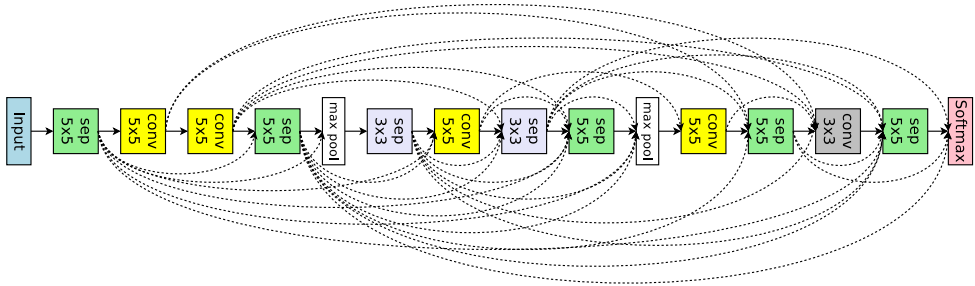}
  \caption{\label{fig:enas_macro}\enas's discovered network from the macro search space for image classification.}
\end{figure*}

\begin{figure}[htb!]
  \centering
  \includegraphics[width=0.45\textwidth]{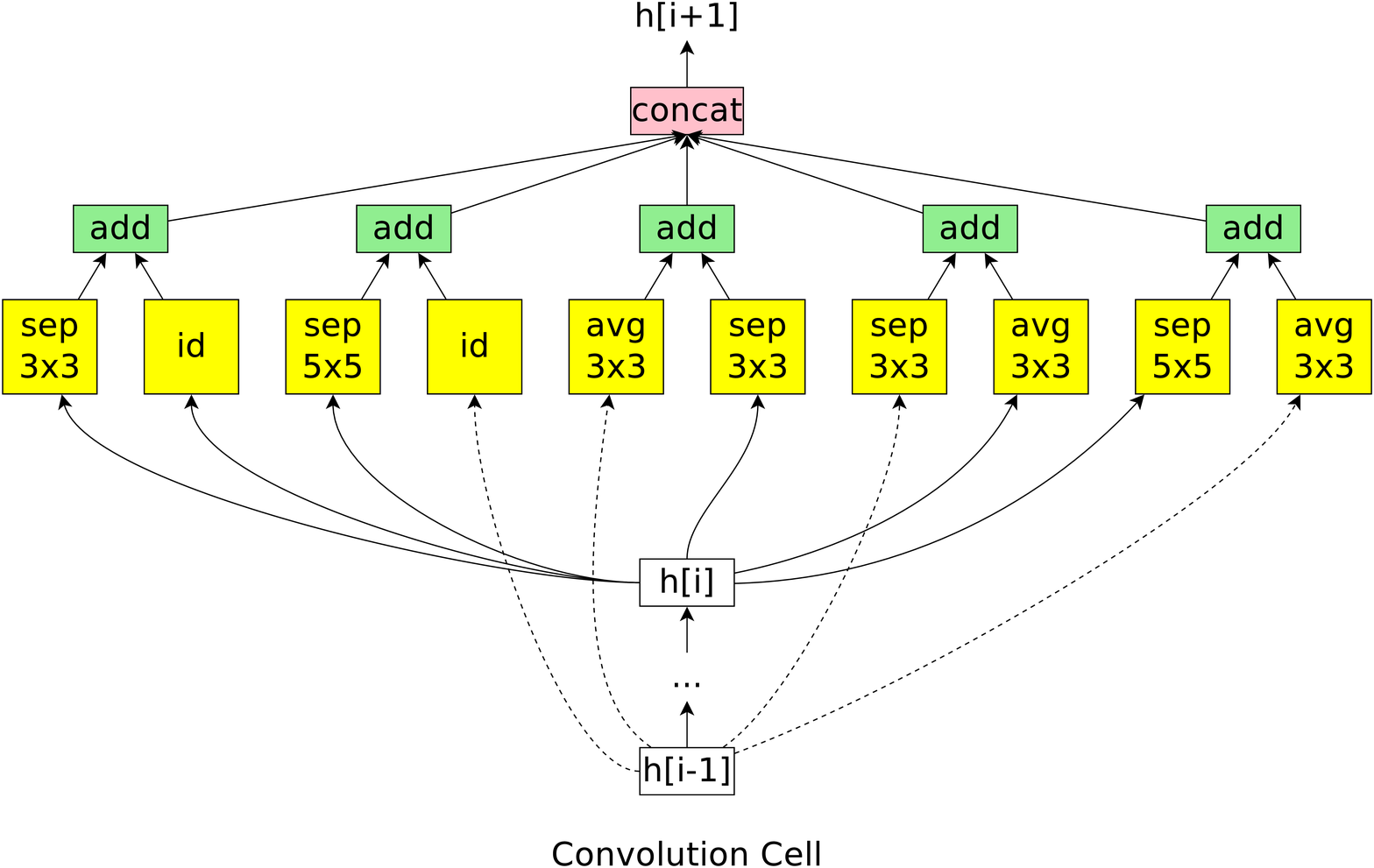} \\~\\
  \includegraphics[width=0.45\textwidth]{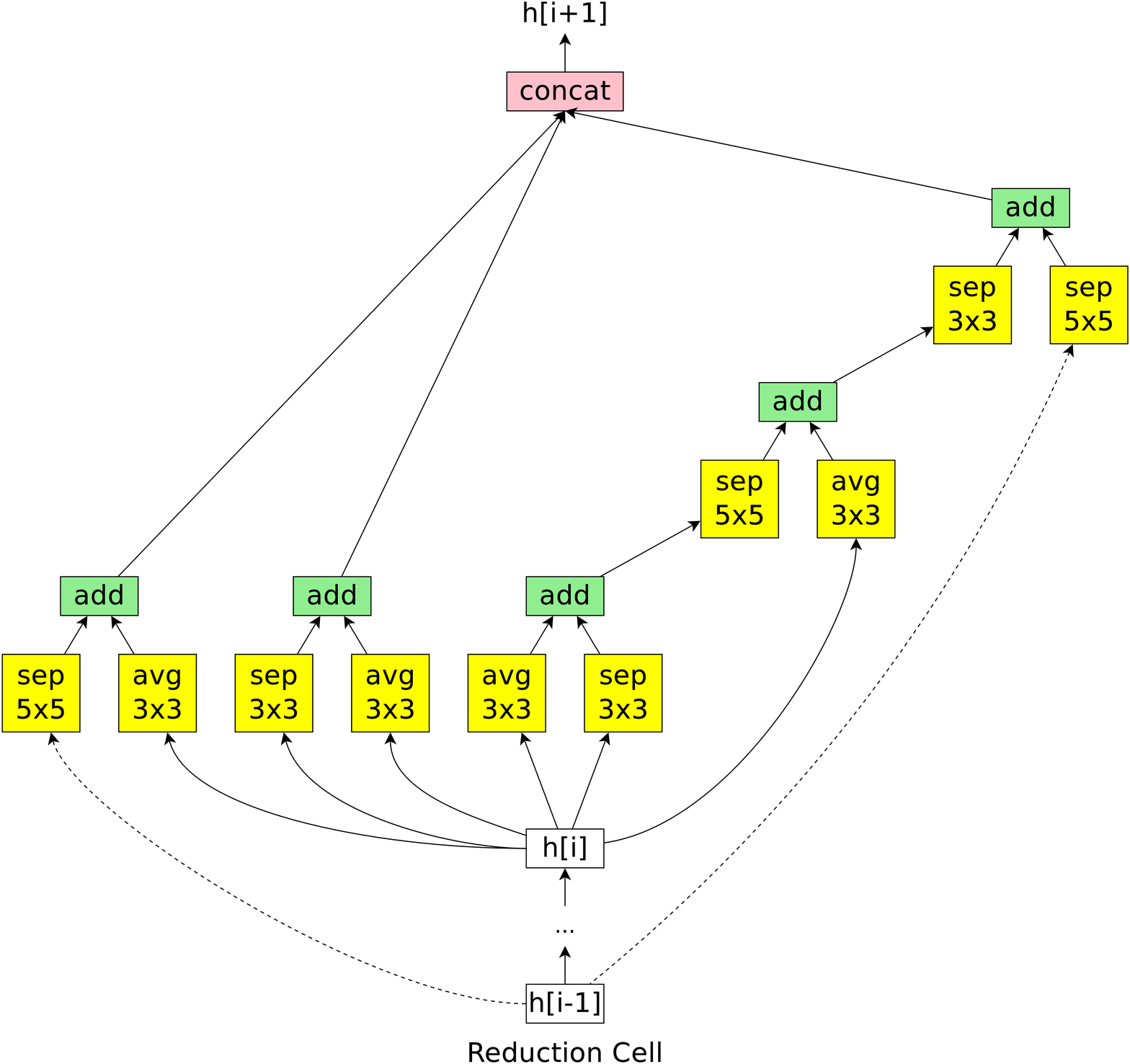}
  \caption{\label{fig:enas_micro}\enas~cells discovered in the micro search space.}
\end{figure}

\paragraph{Results.} Table~\ref{tab:cifar10_results} summarizes the test errors of \enas~and other approaches. In this table, the first block presents the results of DenseNet~\citep{dense_net}, one of the highest-performing architectures that are designed by human experts. When trained with a strong regularization technique, such as Shake-Shake~\citep{shake_shake}, and a data augmentation technique, such as CutOut~\citep{cut_out}, DenseNet impressively achieves the test error of $2.56\%$.

The second block of Table~\ref{tab:cifar10_results} presents the performances of approaches that attempt to design an entire convolutional network, along with the the number of GPUs and the time these methods take to discover their final models. As shown, \enas~finds a network architecture, which we visualize in Figure~\ref{fig:enas_macro}, and which achieves $4.23\%$ test error. This test error is better than the error of $4.47\%$, achieved by the second best NAS model~\citep{nas}. If we keep the architecture, but increase the number of filters in the network's highest layer to $512$, then the test error decreases to $3.87\%$, which is not far away from NAS's best model, whose test error is $3.65\%$. Impressively, \enas~takes about $7$ hours to find this architecture, reducing the number of GPU-hours by more than 50,000x compared to NAS.

The third block of Table~\ref{tab:cifar10_results} presents the performances of approaches that attempt to design one more more modules and then connect them together to form the final networks. \enas~takes $11.5$ hours to discover the convolution cell and the reduction cell, which are visualized in Figure~\ref{fig:enas_micro}. With the convolutional cell replicated for $N=6$ times (\cf~Figure~\ref{fig:micro_search_space}), \enas~achieves $3.54\%$ test error, on par with the $3.41\%$ error of NASNet-A~\citep{nas_module}. With CutOut~\citep{cut_out}, \enas's error decreases to $2.89\%$, compared to $2.65\%$ by NASNet-A.

In addition to \enas's strong performance, we also find that the models found by \enas~are, in a sense, the local minimums in their search spaces. In particular, in the model that \enas~finds from the marco search space, if we replace all separable convolutions with normal convolutions, and then adjust the model size so that the number of parameters stay the same, then the test error increases by $1.7\%$. Similarly, if we randomly change several connections in the cells that \enas~finds in the micro search space, the test error increases by $2.1\%$. This behavior is also observed when \enas~searches for recurrent cells (\cf~Section~\ref{sec:ptb}), as well as in~\citet{nas}. We thus believe that the controller RNN learned by \enas~is as good as the controller RNN learned by NAS, and that the performance gap between NAS and \enas~is due to the fact that we do not sample multiple architectures from our trained controller, train them, and then select the best architecture on the validation data. This extra step benefits NAS's performance.

\subsection{The Importance of \enas}
A question regarding \enas's importance is whether \enas~is actually capable of finding good architectures, or if it is the design of the search spaces that leads to \enas's strong empirical performance.

\paragraph{Comparing to Guided Random Search.} We uniformly sample a recurrent cell, an entire convolutional network, and a pair of convolutional and reduction cells from their search spaces and train them to convergence using the same settings as the architectures found by \enas. For the macro space over entire networks, we sample the skip connections with an activation probability of $0.4$, effectively balancing \enas's advantage from the KL divergence term in its reward (see Section~\ref{sec:cifar10}). Our random recurrent cell achieves the test perplexity of $81.2$ on Penn Treebank, which is far worse than \enas's perplexity of $55.8$. Our random convolutional network reaches $5.86\%$ test error, and our two random cells reache $6.77\%$ on CIFAR-10, while \enas~achieves $4.23\%$ and $3.54\%$, respectively.

\paragraph{Disabling \enas~Search.} In addition to random search, we attempt to train only the shared parameters $\omega$ without updating the controller. We conduct this study for our macro search space (Section~\ref{sec:cifar10_macro}), where the effect of an untrained random controller is similar to dropout with a rate of $0.5$ on the skip connections, and to drop-path on the operations~\citep{nas_module,fractal_net}. At convergence, the model has the error rate of $8.92\%$. On the validation set, an ensemble of $250$ Monte Carlo configurations of this trained model can only reach $5.49\%$ test error. We therefore conclude that the appropriate training of the \enas~controller is crucial for good performance.

\section{\label{sec:related}Related Work and Discussions}
There is a growing interest in improving the efficiency of NAS. Concurrent to our work are the promising ideas of using performance prediction~\citep{nas_performance_pred,nas_peephole}, using iterative search method for architectures of growing complexity~\citep{pnas}, and using hierarchical representation of architectures~\citep{hierarchical_nas}. Table~\ref{tab:cifar10_results} shows that \enas~is significantly more efficient than these other methods, in GPU hours.

\enas's design of sharing weights between architectures is inspired by the concept of weight inheritance in neural model evolution~\citep{nas_evolution,nas_reg_evolution}. Additionally, \enas's choice of representing computations using a DAG is inspired by the concept of stochastic computational graph~\citep{stochastic_computation}, which introduces nodes with stochastic outputs into a computational graph. \enas's utilizes such stochastic decisions in a network to make discrete architectural decisions that govern subsequent computations in the network, trains the decision maker, \ie~the controller, and finally harvests the decisions to derive architectures.

Closely related to \enas~is SMASH~\citep{smash_net}, which designs an architecture and then uses a hypernetwork~\citep{hyper_net} to generate its weight. Such usage of the hypernetwork in SMASH inherently restricts the weights of SMASH's child architectures to a low-rank space. This is because the hypernetwork generates weights for SMASH's child models via tensor products~\citep{hyper_net}, which suffer from a low-rank restriction as for arbitrary matrices $\mathbf{A}$ and $\mathbf{B}$, one always has the inequality: $\text{rank}(\mathbf{A} \cdot \mathbf{B}) \leq \min{\{\text{rank}(\mathbf{A}), \text{rank}(\mathbf{B})\}}$. Due to this limit, SMASH will find architectures that perform well in the restricted low-rank space of their weights, rather than architectures that perform well in the normal training setups, where the weights are no longer restricted. Meanwhile, \enas~allows the weights of its child models to be arbitrary, effectively avoiding such restriction. We suspect this is the reason behind \enas's superior empirical performance to SMASH. In addition, it can be seen from our experiments that \enas~can be flexibly applied to multiple search spaces and disparate domains, \eg~the space of RNN cells for the text domain, the macro search space of entire networks, and the micro search space of convolutional cells for the image domain.

\section{\label{sec:conclusion}Conclusion}
NAS is an important advance that automatizes the designing process of neural networks. However, NAS's computational expense prevents it from being widely adopted. In this paper, we presented \enas, a novel method that speeds up NAS by more than 1000x, in terms of GPU hours. \enas's key contribution is the sharing of parameters across child models during the search for architectures. This insight is implemented by searching for a subgraph within a larger graph that incorporates architectures in a search space. We showed that \enas~works well on both CIFAR-10 and Penn Treebank datasets.

\section*{Acknowledgements}
The authors want to thank Jaime Carbonell, Zihang Dai, Lukasz Kaiser, Azalia Mirhoseini, Ashwin Paranjape, Daniel Selsam, and Xinyi Wang for their suggestions on improving the paper.

\bibliography{main}
\bibliographystyle{icml2018}


\newpage
\twocolumn[
\icmltitle{Appendix for:\\Efficient Neural Architecture Search via Parameter Sharing}
]
\appendix
\section{\label{sec:ptb_details}Details on Penn Treebank Experiments}

\paragraph{Computations in an RNN Cell.}~We think of the cell at time step $t$ as a DAG with $N$ computational nodes, indexed by $\h^{(t)}_1$, $\h^{(t)}_2$, ... $\h^{(t)}_N$. Node $\h^{(t)}_1$ receives two inputs: 1) the RNN signal $\x^{(t)}$ at its current time step; and 2) the output $\h^{(t-1)}_D$ from the cell at the previous time step. The following computations are performed:
\begin{align}
  \label{eqn:rhn_skip}
  \c^{(t)}_1 &\leftarrow \text{sigmoid}\left(
    \x^{(t)} \cdot \mathbf{W}^{(\x, \c)} + \h^{(t-1)}_{N} \cdot, \mathbf{W}^{(\c)}_0
  \right) \\
  \h^{(t)}_1 &\leftarrow
    \c^{(t)}_1 \otimes f_1\left( \x^{(t)} \cdot \mathbf{W}^{(\x, \h)} + \h^{(t-1)}_{N} \cdot \mathbf{W}^{(\h)}_1 \right) \nonumber \\
    &+ (1 - \c^{(t)}_1) \otimes \h^{(t-1)}_N,
\end{align}
where $f_1$ is an activation function that the controller will decide. For $\ell = 2, 3, ..., N$, node $\h_\ell$ receives its input from a layer $j_{\ell} \in \{\h_1, ..., \h_{\ell-1}\}$, which is specified by the controller, and then performs the following computations:
\begin{align}
  \label{eqn:rhn_skip}
  \c^{(t)}_\ell &\leftarrow \text{sigmoid}\left(
    \h^{(t)}_{j_\ell} \cdot \mathbf{W}^{(\c)}_{\ell, j_\ell}
  \right) \\
  \h^{(t)}_\ell &\leftarrow
    \c^{(t)}_\ell \otimes f_\ell \left(
    \h^{(t)}_{j_\ell} \cdot \mathbf{W}^{(\h)}_{\ell, j_\ell} \right)
    + (1 - \c^{(t)}_\ell) \otimes \h^{(t)}_{j_\ell}.
\end{align}
Therefore, the shared parameters $\mathbf{\omega}$ among different recurrent cells consist of all the matrices $\mathbf{W}^{(\x, \c)}$, $\mathbf{W}^{(\x, \h)}$, $\mathbf{W}^{(\c)}_{\ell, j}$, $\mathbf{W}^{(\h)}_{\ell, j}$, word embeddings, and the softmax weights if they are not tied with the word embeddings. The controller decides the connection $j_\ell$ and the activation function $f_\ell$ for each $\ell \in \{2, 3, ..., N\}$. The layers that are never selected by any subsequent layers are averaged and sent to a softmax head, or to higher recurrent layers.

\paragraph{Parameters Initialization.}~Our controller's parameters $\theta$ are initialized uniformly in $[-0.1, 0.1]$. We find that for Penn Treebank, \enas~quite insensitive to its initialization than for CIFAR-10. Meanwhile, the shared parameters $\omega$ are initialized uniformly in $[-0.025, 0.025]$ during architecture search, and $[-0.04, 0.04]$ when we train a fixed architecture recommended by the controller.

\paragraph{Stabilizing the Updates of $\omega$.}~To stabilize the updates of $\mathbf{\omega}$, during the architectures search phase, a layer of batch normalization~\citep{batch_norm} is added immediately after the average of these layers, before the average are sent out of the cell as its output. When a fixed cell is sampled by the controller, we find that we can remove the batch normalization layer without any loss in performance.

\section{\label{sec:cifar10_details}Details on CIFAR-10 Experiments}
We find the following tricks crucial for achieving good performance with \enas. Standard NAS~\cite{nas,nas_module} rely on these and other tricks as well.

\paragraph{Structure of Convolutional Layers.}~Each convolution in our model is applied in the order of relu-conv-batchnorm~\citep{batch_norm,skip_connections_order}. Additionally, in our micro search space, each depthwise separable convolution is applied twice~\citep{nas_module}.

\paragraph{Stabilizing Stochastic Skip Connections.}~If a layer receives skip connections from multiple layers before it, then these layers' outputs are concatenated in their depth dimension, and then a convolution of filter size $1 \times 1$ (followed by a batch normalization layer and a ReLU layer) is performed to ensure that the number of output channels does not change between different architectures. When a fixed architecture is sampled, we find that one can remove these batch normalization layers to save computing time and parameters of the final model, without sacrificing significant performance.
  
\paragraph{Global Average Pooling.}~After the final convolutional layer, we average all the activations of each channel and then pass them to the Softmax layer. This trick was introduced by~\cite{network_in_network}, with the purpose of reducing the number of parameters in the dense connection to the Softmax layer to avoid overfitting.

The last two tricks are extremely important, since the gradient updates of the shared parameters $\omega$, as described in Eqn~\ref{eqn:update_omega}, have very high variance. In fact, we find that without these two tricks, the training of \enas~is very unstable.

\end{document}